# Damage-Aware Bandit Pruning for Vision and Language Transformers

**Salem Ameen and Sunil Vadera**

School of Science, Engineering and Environment, University of Salford, Salford, United Kingdom

Corresponding author: Salem Ameen (s.a.ameen1@salford.ac.uk)

## Abstract

Structured post-training pruning of transformers requires selecting complete functional units whose suppression causes limited degradation. We formulate structured-unit selection for language and vision transformers as a damage-aware multi-armed bandit problem under a fixed candidate-evaluation budget. Attention heads and MLP channel groups are temporarily masked on calibration batches. Paired damage is the masked loss minus the base loss on the same batch, reducing batch-to-batch variation. A smooth bounded reward drives either a UCB-style policy or fractional-Beta Thompson Sampling, and the final mask is constructed sequentially by adding one unit at each step. The selected units are functionally zeroed in the original dense checkpoint; therefore, the reported parameter effects represent effective structural suppression rather than physical compression or measured speedup. Experiments on WikiText-2, LAMBADA, and Imagenette cover GPT-2, OPT, Pythia, Qwen2.5, SmolLM2, ViT-B/16, DeiT-Tiny, and Swin-Tiny, with comparisons against random, magnitude, static-saliency, and budgeted-greedy selection. Across five seeds, the bandit methods usually reduce degradation relative to budgeted greedy in the paired language-model comparisons. Of 28 comparisons highlighted in the paper, 23 bootstrap confidence intervals exclude zero and 11 paired tests have $p < 0.05$; six have $q < 0.05$ after Benjamini-Hochberg correction across the full family of 116 dataset-wise tests. Matched-evaluation results for ViT-B/16 and Swin-Tiny indicate that their gains are not explained solely by a larger candidate-evaluation budget.

**Keywords**: transformer pruning; structured-unit masking; functional zeroing; attention-head pruning; multi-armed bandits; paired damage estimation.

## 1. Introduction

Neural-network pruning has evolved from curvature-based parameter saliency [1,2] to large-scale connection pruning [3]. Transformers introduce structured candidates, particularly attention heads and feed-forward channels, whose suppression can be related more directly to complete computational units. Prior analyses show that head importance is highly uneven and that many heads can be removed with limited degradation [4,5]. The practical difficulty is to identify removable units without exhaustively testing every candidate on large calibration sets.

This work addresses selection as a fixed-budget multi-armed bandit problem. At each pull, one candidate attention head or MLP group is temporarily masked, and its damage is measured as the change in loss relative to the current masked model on the same mini-batch. The paired observation is converted into a bounded reward. MAB-UCB and MAB-TS, based respectively on upper-confidence exploration [6] and Thompson sampling [7], allocate repeated evaluations among candidates, while the final top-k mask is built one unit at a time. This sequential conditioning accounts for units already selected but does not solve the joint combinatorial pruning problem.

The study evaluates the same selection principle on decoder-only language models and vision transformers using WikiText-2, LAMBADA, and Imagenette. The selected units are functionally zeroed in the original dense checkpoint; no physical tensors are removed. Accordingly, the experiments assess structured-unit selection quality and effective suppression, not checkpoint compression, memory reduction, or deployment speedup.

**The paper makes five contributions:**

- A damage-aware MAB formulation for post-training transformer structured-unit selection, where attention heads and MLP groups are candidate arms.
- A paired damage estimator that evaluates the base and temporarily masked model on the same calibration batch, reducing batch-to-batch noise in unit comparison.
- A bounded reward based on temperature-scaled damage, used by both MAB-UCB and fractional-Beta MAB-TS (Thompson Sampling).

- An empirical comparison against random, magnitude, Wanda-style, Taylor, Fisher, and budgeted greedy baselines.
- A five-seed evaluation on WikiText-2, LAMBADA, and Imagenette, including GPT-style language models and ViT, DeiT, and Swin vision transformers, while explicitly distinguishing functional zeroing from physical checkpoint compression.

## 2. Related work and positioning

Early studies of transformer redundancy examined attention heads directly. Michel et al. [4] showed that many heads can be removed at test time with limited effect, and Voita et al. [5] linked resistance to pruning with specialized linguistic roles. These studies establish head-level redundancy but do not address how a limited evaluation budget should be allocated when candidate effects are noisy.

Transformer-pruning methods span different structures and objectives. SparseGPT [8] and Wanda [9] target one-shot weight sparsity in large language models, whereas the fast post-training framework of Kwon et al. [10] searches structured attention-head and feed-forward masks under FLOP or latency constraints. LLM-Pruner [11] removes coupled LLM structures using gradient information and applies lightweight recovery tuning. FLAP [12] provides retraining-free structured pruning based on activation fluctuations, while SliceGPT [13] reduces dense matrix dimensions by deleting rows and columns. For vision transformers, X-Pruner [14] learns explainability-aware structured masks. These approaches differ in their use of static saliency, gradients or activations, physical model reduction, and recovery training.

The authors' previous work formulated scalar-weight pruning as fixed-budget MAB selection [15] and subsequently applied loss-aware bandit search to structured neurons [16] and convolutional feature maps [17]. Those methods distinguish temporary evaluation from final pruning and provide the immediate methodological foundation for the present study.

The present method differs in three respects. First, the candidate arms are transformer attention heads or contiguous MLP channel groups. Second, each reward is derived from paired same-batch damage under the units already selected. Third, the mask is constructed sequentially under an explicit run-specific evaluation budget. Unlike methods that physically reduce model structures [10-14], the current implementation leaves the dense checkpoint intact and uses functional zeroing. The contribution is therefore a budgeted selection rule rather than a claim of realised compression or acceleration.

## 3. Method

This section defines the damage-aware structured-unit selection method. The formulation retains the fixed-budget arm-selection principle of the authors' previous MAB pruning work [15-17] and adapts it to typed transformer units, paired loss measurements, and sequential mask construction.

Figure 1 shows the implemented pipeline and separates adaptive unit selection from the final evaluation under functional zeroing.

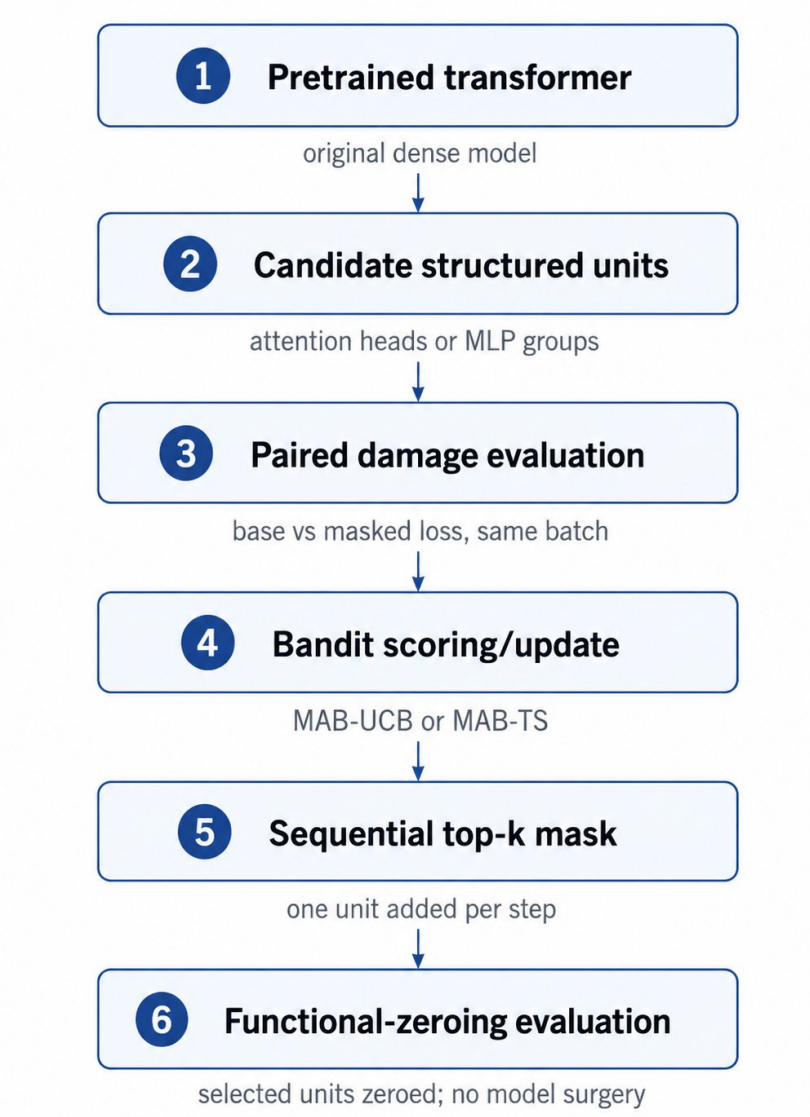

*Figure 1. Damage-aware bandit structured-unit selection pipeline. Candidate attention heads or MLP groups are temporarily masked and evaluated by paired damage on calibration batches. A bandit policy scores candidates and constructs a sequential top-k mask, adding one unit at each step. The original dense model is then evaluated with the selected units functionally zeroed. The pipeline measures structured-unit selection quality and does not imply physical checkpoint compression, memory reduction, or measured deployment speed-up.*

## 3.1 Candidate structured units

Let $f_\theta$ be a pretrained transformer with parameter set $\theta$, and let $\mathscr{L}_p$ denote the set of layers eligible for pruning. The candidate space is typed. The label $H$ denotes an attention-head candidate, and the label $M$ denotes an MLP-group candidate. These labels make the two candidate families disjoint even when their layer and position indices coincide.

The attention-head candidate set is

$$U_H = \{(H,l,h) : l \in \mathscr{L}_p,\ h = 1,\ldots,H_l\}. \tag{1}$$

Here $H_l$ is the number of attention heads in layer $l$. For MLP pruning, let $D_l$ be the expanded hidden dimension of the MLP block in layer $l$. The expanded channels are partitioned into groups $G_{l,q}$ of nominal width $g_0$, except that the final group may be smaller. The number of groups and the width of group $q$ are

$$Q_l = \text{ceil}\left(\frac{D_l}{g_0}\right), \qquad w_{l,q} = \left|G_{l,q}\right|. \tag{2}$$

The MLP-group candidate set is

$$U_M = \{(M,l,q) : l \in \mathscr{L}_p,\ q = 1,\ldots,Q_l\}. \tag{3}$$

The full candidate set depends on the enabled pruning target:

$$U = \begin{cases} U_H, & \text{attention - head pruning only,} \\ U_M, & \text{MLP - group pruning only,} \\ U_H \cup U_M, & \text{both target types enabled.} \end{cases} \tag{4}$$

The number of available candidates is

$$K = |U|. \tag{5}$$

When both target types are enabled, $U_H \cap U_M = \varnothing$ because of the type labels, and therefore $K = \left|U_H\right| + \left|U_M\right|$. Given a pruning ratio $\rho$, the target number of selected units is

$$k = \max\{1, \text{round}(\rho K)\}. \tag{6}$$

The selected set is constructed sequentially. At selection step $s$, the previously selected set is $S_{s-1}$, and the remaining candidate pool $R_s$ and active pool $A_s$ satisfy

$$R_s = \{u \in U : u \notin S_{s-1}\}, \qquad A_s \subseteq R_s. \tag{7}$$

If no active-pool restriction is used, then $A_s = R_s$. Thus, $A_s$ is a computational subset of the remaining candidates, not a new pruning target.

### 3.2 Structured-unit masks

The masks are defined on the same typed units as the candidate set. For a selected pruning set $S \subseteq U$, the induced binary masks are

$$m_{l,h}^{H}(S) = \begin{cases} 0, & (H,l,h) \in S, \\ 1, & (H,l,h) \notin S, \end{cases} \qquad m_{l,q}^{M}(S) = \begin{cases} 0, & (M,l,q) \in S, \\ 1, & (M,l,q) \notin S. \end{cases} \tag{8}$$

For an attention-head candidate $(H,l,h)$, let $H_{l,h}(x)$ denote the output of head $h$ in layer $l$ before head concatenation and output projection. Functional zeroing applies the head mask to the whole head output:

$$\tilde{H}_{l,h}(x;S) = m_{l,h}^{H}(S)\, H_{l,h}(x). \tag{9}$$

For an MLP-group candidate $(M,l,q)$, let $z_l(x) \in \mathbb{R}^{D_l}$ be the expanded MLP activation after the first projection and nonlinearity. Each expanded channel $r$ belongs to exactly one group, denoted $G_{l,q(r)}$. The masked activation is

$$\tilde{z}_{l,r}(x;S) = m_{l,q(r)}^{M}(S)\, z_{l,r}(x), \qquad r = 1,\ldots,D_l. \tag{10}$$

This definition leaves all channels in unselected groups unchanged. Hence, an attention-head mask removes a complete attention subspace, whereas an MLP-group mask removes a contiguous block of feed-forward hidden channels.

The masked transformer evaluated after selecting $S$ is written as

$$f_{\theta,S}(x) = f_{\theta}\left(x; m^{H}(S), m^{M}(S)\right), \qquad f_{\theta,\varnothing}(x) = f_{\theta}(x). \tag{11}$$

Only the relevant mask family is used in single-target experiments. For example, an attention-head-only run uses $m^{H}(S)$ and leaves the MLP mask equal to one.

### 3.3 Functional zeroing and parameter accounting

The experiments use functional zeroing rather than physical dense-model rewriting. Functional zeroing suppresses the forward contribution of selected units, but the dense checkpoint tensors remain stored. Consequently, the reported parameter reduction is an effective zeroed-parameter estimate, not a measured checkpoint-size reduction.

For a standard multi-head attention layer with model dimension $d$ and head dimension $d_h$, the effective parameter contribution of one independently removable head is approximated by

$$P_h^{\text{MHA}} \approx 4dd_h, \qquad P_h^{\text{MHA+bias}} \approx 4dd_h + 3d_h. \tag{12}$$

The first approximation accounts for the head-specific query, key, value, and output-projection slices. The second also includes head-specific bias terms when those terms are explicitly zeroed. For grouped-query attention, query heads may share key-value heads, so a query-head mask should not automatically count a shared key-value head as removed. The query/output-only approximation is

$$P_h^{\text{GQA}} \approx 2dd_h. \tag{13}$$

For an MLP group in layer $l$ with width $w_{l,q}$, the non-gated and gated approximations are

$$P_m^{\text{non-gated}}(l,q) \approx 2dw_{l,q} + w_{l,q}, \qquad P_m^{\text{gated}}(l,q) \approx 3dw_{l,q} + 2w_{l,q}. \tag{14}$$

For the final selected set $S_k$, the effective zeroed-parameter percentage is

$$Z\left(S_k\right) = 100 \times \frac{\sum_{j \in S_k} P_j}{P_{\text{total}}}. \tag{15}$$

Here $P_j$ is the architecture-specific effective contribution of selected unit $j$, and $P_{\text{total}}$ is the nominal parameter count of the original dense checkpoint.

### 3.4 Paired damage estimation

At selection step $s$, the units in $S_{s-1}$ are treated as already suppressed while a new candidate is tested. For a calibration mini-batch $B$, the base loss under the current selected mask is

$$L_0\left(B, S_{s-1}\right) = \frac{1}{|B|} \sum_{\left(x_i, y_i\right) \in B} \ell\left(f_{\theta, S_{s-1}}\left(x_i\right), y_i\right). \tag{16}$$

If candidate $j \in A_s$ is temporarily added to the current mask, the paired masked loss is

$$L_j\left(B, S_{s-1}\right) = \frac{1}{|B|} \sum_{\left(x_i, y_i\right) \in B} \ell\left(f_{\theta, S_{s-1} \cup \{j\}}\left(x_i\right), y_i\right). \tag{17}$$

The paired damage of candidate $j$ on the same mini-batch is

$$d_j\left(B, S_{s-1}\right) = L_j\left(B, S_{s-1}\right) - L_0\left(B, S_{s-1}\right). \tag{18}$$

This paired construction compares the masked and unmasked conditions on exactly the same examples. Positive damage means that temporarily removing candidate $j$ increases the loss; near-zero damage means that the candidate has little measured effect; and negative damage means that the temporary mask improves the loss on that mini-batch. The estimate is marginal and conditioned on $S_{s-1}$. Interactions among the finally selected units are assessed by evaluating $f_{\theta, S_k}$ after the whole set has been constructed. Thus, each new candidate is assessed in the context of the units already selected, but the method does not jointly search all multi-unit combinations.

### 3.5 Smooth bounded reward

The paired damage is converted into a bounded safe-removal reward with temperature $\tau > 0$:

$$r_j\left(B, S_{s-1}\right) = \frac{1}{1 + \exp\{d_j\left(B, S_{s-1}\right)/\tau\}}. \tag{19}$$

The reward lies between zero and one. A candidate with zero measured damage receives reward 0.5. A harmful candidate, whose removal increases the loss, receives reward below 0.5. A candidate whose temporary removal reduces the loss receives reward above 0.5. In the implementation, the exponential argument is clipped only for numerical stability; clipping does not change the monotonic relation between damage and reward.

### 3.6 MAB-UCB selection and update

At selection step $s$, MAB-UCB is run over the active pool $A_s$ for $T_{\text{step}}$ pulls. For candidate $j$, let $n_j(t)$ be the number of times it has been evaluated before pull $t$, and let $\mu_j(t)$ be its empirical mean reward. After observing reward $r_j(t)$, the statistics are updated by

$$n_j(t+1) = n_j(t) + 1, \qquad \mu_j(t+1) = \mu_j(t) + \frac{r_j(t) - \mu_j(t)}{n_j(t+1)}. \tag{20}$$

Let the candidates already evaluated before pull $t$ and the total number of completed pulls be

$$E_s(t) = \{j \in A_s : n_j(t) > 0\}, \qquad N_t = \sum_{j \in E_s(t)} n_j(t). \tag{21}$$

Unevaluated candidates are sampled first, until each active candidate has been tried once or the pull budget is exhausted. Once at least one candidate has been evaluated, the UCB rule selects

$$j_t = \arg\max_{j \in E_s(t)} \left[\mu_j(t) + c_{\text{ucb}} \sqrt{\frac{\log N_t}{n_j(t)}}\right]. \tag{22}$$

After $T_{\text{step}}$ pulls, the candidate selected for permanent inclusion in the pruning set is

$$j_s^{\mathrm{UCB}} = \arg \max_{j \in E_s\left(T_{\mathrm{step}}\right)} \mu_j\left(T_{\mathrm{step}}\right). \tag{23}$$

The constant $c_{\mathrm{ucb}}$ controls the exploration bonus. Because this constant is configurable, the method is described as MAB-UCB rather than as the fixed-constant UCB1 rule.

### 3.7 Fractional-Beta MAB-TS

MAB-TS uses the same bounded rewards but maintains Beta pseudo-counts for candidates in the active pool. At the beginning of each selection step,

$$\alpha_j(0) = 1, \qquad \beta_j(0) = 1, \qquad j \in A_s. \tag{24}$$

At pull $t$, a Thompson sample is drawn for each active candidate and the largest sample is selected:

$$\vartheta_j(t) \sim \mathrm{Beta}\left(\alpha_j(t), \beta_j(t)\right), \qquad j_t = \operatorname*{argmax}_{j \in A_s} \vartheta_j(t). \tag{25}$$

After observing bounded reward $r_j(t)$, the pseudo-counts for the sampled candidate are updated by

$$\alpha_j(t+1) = \alpha_j(t) + r_j(t), \qquad \beta_j(t+1) = \beta_j(t) + 1 - r_j(t). \tag{26}$$

This is a fractional-Beta pseudo-count mechanism for bounded rewards, not an exact conjugate posterior for Bernoulli observations. In parallel, the implementation keeps $n_j(t)$ and $\mu_j(t)$ using the empirical-mean update equations in Section 3.6. The candidate finally selected at step $s$ is

$$j_s^{\mathrm{TS}} = \arg \max_{j \in E_s\left(T_{\mathrm{step}}\right)} \mu_j\left(T_{\mathrm{step}}\right). \tag{27}$$

The posterior mean $\alpha_j / \left(\alpha_j + \beta_j\right)$ is retained as an uncertainty-smoothed diagnostic, while the implemented final selection within a step uses the observed empirical mean reward.

### 3.8 Sequential top-k mask construction and cost

The complete method is a sequential top-k mask-construction procedure. It does not delete several candidates simultaneously. Instead, it repeats the same selection step until $k$ units have been chosen.

**Algorithm 1. Damage-aware bandit structured-unit selection.** The inputs are the pretrained transformer, candidate set $U$, pruning ratio $\rho$, active-pool rule, per-step pull budget $T_{\mathrm{step}}$, calibration mini-batch sampler, reward temperature $\tau$, and the chosen bandit policy. The output is the selected structured-unit set $S_k$ and the final evaluation under functional zeroing.

| |
|---|
| *Step 1: Initialise $S_0 = \varnothing$ and $R_1 = U$.* |
| *Step 2: For selection step $s = 1, \ldots, k$, choose an active pool $A_s \subseteq R_s$.* |
| *Step 3: Run the chosen bandit policy on $A_s$. Each pull samples a calibration mini-batch, evaluates the current base loss, temporarily adds one candidate to the mask, evaluates the paired masked loss, converts damage to reward, and updates only that candidate's statistics.* |
| *Step 4: After the pull budget is exhausted, select the evaluated candidate with the largest step-level safe-removal score.* |
| *Step 5: Add the selected candidate to the cumulative set and remove it from the remaining pool.* |
| *Step 6: After all k selections, evaluate the dense checkpoint under the final functional mask $S_k$.* |

The number of paired candidate evaluations used by the MAB search is

$$E_{\mathrm{MAB}} = k T_{\mathrm{step}}. \tag{28}$$

If each paired evaluation uses $b_{\text{pull}}$ calibration mini-batches and each mini-batch forward pass has cost $C_F$, the dominant search cost is approximately

$$C_{\text{search}} \approx 2 b_{\text{pull}} C_F k T_{\text{step}}. \tag{29}$$

The policy-selection overhead is

$$C_{\text{policy}} = O\left(\sum_{s=1}^{k} T_{\text{step}} \left|A_s\right|\right). \tag{30}$$

A budgeted greedy baseline that evaluates $G$ candidates per step uses

$$E_{\text{greedy}} = kG. \tag{31}$$

Thus, greedy and MAB can be compared at the same paired candidate-evaluation budget by setting $G = T_{\text{step}}$.

### 3.9 Baseline scoring functions

All baselines operate on the same candidate set $U$ and select $k$ units. Random selection samples uniformly among all $k$-element subsets:

$$S_k \sim \text{Unif}\{S \subseteq U : |S| = k\}. \tag{32}$$

Magnitude pruning scores candidate $j$ by the average absolute value of its associated parameters $W_j$:

$$a_j^{\text{mag}} = \frac{1}{\left|W_j\right|} \sum_{w \in W_j} |w|. \tag{33}$$

Following the weight-and-activation principle introduced by Sun et al. [9], the structured Wanda-style baseline combines parameter magnitude with activation scale:

$$a_j^{\text{Wanda}} = \sum_{(a,b) \in W_j} \left|w_{ab}\right| \; \| X_b \|_2. \tag{34}$$

Taylor saliency uses first-order loss sensitivity,

$$a_j^{\text{Taylor}} = \left| \sum_{w \in W_j} w \frac{\partial \mathcal{L}}{\partial w} \right|, \tag{35}$$

and the Fisher-style score uses the squared gradient-weight product:

$$a_j^{\text{Fisher}} = \sum_{w \in W_j} \left( w \frac{\partial \mathcal{L}}{\partial w} \right)^2. \tag{36}$$

Budgeted greedy selection directly evaluates candidate damage within a per-step trial set $G_s \subseteq R_s$:

$$j_s^{\text{greedy}} = \underset{j \in G_s}{\text{argmax}}\, r_j\left(B, S_{s-1}\right). \tag{37}$$

The greedy baseline is loss-aware, but it performs direct candidate comparison rather than an exploration-exploitation policy. When $\left|G_s\right| = T_{\text{step}}$, the comparison isolates the selection rule rather than the evaluation budget.

## 4. Experimental setup

Unless otherwise stated, stochastic results are reported as mean ± standard deviation over five seeds: 1, 2, 3, 42, and 123. For language modelling, lower perplexity and lower percent perplexity increase are better; negative percent change means that the pruned/masked model obtains lower perplexity than the unmasked reference under the same evaluation protocol. For vision transformers, lower cross-entropy loss is better, while higher Top-1 and Top-5 accuracy are better. Negative delta loss indicates improved loss after pruning; negative delta Top-1 or Top-5 indicates accuracy degradation. In all result tables, evals denotes adaptive paired candidate-evaluation trials or arm pulls used during selection. Because the proposed damage estimator is paired, one such trial requires one base forward evaluation and one masked forward evaluation on the same calibration batch; the number of calibration forward passes is therefore approximately twice the reported paired-trial count. Static saliency baselines such as magnitude, Wanda-style, Taylor, and Fisher are deterministic under a fixed model, calibration subset, and implementation configuration; they are therefore reported as single fixed values when repeated seeds would not change the ranking. Random, greedy, and bandit methods are reported over seeds when their selection process or sampled calibration batches vary across runs.

*Table 1. Compared pruning techniques and how calibration use differs from counted paired candidate-evaluation trials.*

| Technique | Calibration use | Counted evals | Role in this study |
|---|---|---|---|
| Random | No | 0 | Uniform random structured-unit selection. |
| Magnitude | No | 0 | Static parameter-magnitude ranking. |
| Wanda-style | Yes: activations | 0 paired trials | Activation-aware weight saliency. |
| Taylor | Yes: gradients | 0 paired trials | First-order loss-sensitivity saliency. |
| Fisher | Yes: gradients | 0 paired trials | Squared-gradient Fisher-style saliency. |
| Budgeted greedy | Yes | Matched paired trials when reported | Directly tests low-damage candidate removals. |
| MAB-UCB | Yes | Paired trials | Damage-aware bandit selection with UCB exploration. |
| MAB-TS | Yes | Paired trials | Damage-aware Thompson-style selection. |

*Table 2. Datasets and metrics. WikiText-2 and LAMBADA are never mixed in the same aggregate table.*

| Dataset | Role in this paper | Configuration | Split/use | Metric |
|---|---|---|---|---|
| WikiText-2 | Main calibration and evaluation dataset | Salesforce/wikitext, wikitext-2-raw-v1 | Calibration: train split; evaluation: held-out split | Perplexity |
| LAMBADA | Cross-dataset robustness evaluation | cimec/lambada, plain_text | Evaluation: test split; calibration remains WikiText-2 | Perplexity |
| Imagenette | Vision-transformer evaluation | imagenette2-320 image-folder | train for calibration, val for evaluation | Cross-entropy loss, Top-1, Top-5 |

The language-model suite comprises GPT-2 [18], OPT [19], Pythia [20], Qwen2.5-0.5B [21], and SmolLM2-360M [22]. WikiText-2 [23] is used for the main language evaluation and calibration, while LAMBADA [24] provides a cross-dataset evaluation. The vision suite comprises ViT-B/16 [25], DeiT-Tiny [26], and Swin-Tiny [27] on Imagenette [28]. These models cover global attention, data-efficient training, and hierarchical shifted-window attention, while SmolLM2 is treated as a compact-model sensitivity case.

The main language-model search configuration uses sequence length 128, up to 512 calibration texts, up to 1024 evaluation texts, and 80 evaluation batches. The vision-transformer runs use Imagenette image-folder data, with training images for calibration and validation images for evaluation. The vision batch size is 8, with up to 1024 calibration images and up to 2000 evaluation images. MAB and budgeted-greedy methods use bounded candidate-evaluation budgets. Static saliency methods do not perform adaptive candidate-selection trials, so their selection-cost count is reported as zero.

For reproducibility, the language-model runs use a maximum pre-screened pool of 48 candidates under the low-magnitude screening rule. Unless otherwise stated, the per-step bandit budget is $T_{\text{step}} = 32$ paired candidate evaluations, each paired damage estimate uses $b_{\text{pull}} = 2$ calibration mini-batches, the budgeted-greedy trial count is $G = 32$, the UCB exploration constant is $c_{\text{ucb}} = 1.5$, the MLP group size is $g_0 = 32$, and the reward temperature is $\tau = 0.02$.

When the active-pool size is not set manually, the active pool is selected from the remaining candidate pool using the seeded rule

$$|A_s| = \min\{|R_s|, \max\left(2, \lfloor 2\sqrt{|R_s|} \rfloor\right)\}.$$

If $T_{\text{step}} < |A_s|$, only the evaluated subset $E_s$ is eligible for final selection at that step. The greedy trial set $G_s$ is selected from the remaining candidates using the same seeded candidate-pool rule. For the vision-transformer runs, the total paired candidate-evaluation budget used by each method is reported explicitly in the Evals column of Tables 8 and 9. The nominal vision pruning ratio is converted to an integer head count by rounding; for DeiT-Tiny, a nominal ratio of 0.10 therefore selects 4 of 36 heads (11.11%).

Wanda-style, Taylor, and Fisher baselines are reported for the main WikiText-2 language experiments. The LAMBADA and Imagenette tables focus on the baselines available under the corresponding cross-dataset and image-folder protocols; unavailable baselines are not counted as wins for any method.

*Table 3. Parameter-accounting table. Values are effective zeroed-parameter estimates under functional zeroing, not measured physical checkpoint compression. DeiT-Tiny has 12 blocks with 3 heads per block (36 heads) [26], so integer rounding at the nominal 0.10 ratio selects 4/36 heads. For Swin-Tiny, selected heads may lie in stages with different channel dimensions, so the effective zeroed-parameter percentage varies across seeds and methods and is reported as a range.*

| Model | Target | Nominal prune ratio | Actual unit ratio | Nominal params before | Effective zeroed params | Approx. params zeroed | Approx. active params after |
|---|---|---|---|---|---|---|---|
| gpt2 | Head | 10% | 9.72% | 124M | 2.2141% | 2.75M | 121.25M |
| gpt2-medium | Head | 10% | 9.90% | 355M | 2.8095% | 9.97M | 345.03M |
| facebook/opt-125m | Head | 10% | 9.72% | 125M | 2.1999% | 2.75M | 122.25M |
| facebook/opt-350m | Head | 10% | 9.90% | 350M | 3.0099% | 10.53M | 339.47M |
| EleutherAI/pythia-160m | Head | 10% | 9.72% | 160M | 1.6974% | 2.72M | 157.28M |
| EleutherAI/pythia-410m | Head | 10% | 9.90% | 410M | 2.4594% | 10.08M | 399.92M |
| Qwen/Qwen2.5-0.5B | Head | 10% | 10.12% | 500M | 0.7893% | 3.95M | 496.05M |
| gpt2 | MLP group | 3% | 3.04% | 124M | 1.3834% | 1.72M | 122.28M |
| gpt2 | MLP group | 5% | 5.03% | 124M | 2.2924% | 2.84M | 121.16M |
| gpt2 | MLP group | 8% | 7.99% | 124M | 3.6362% | 4.51M | 119.49M |
| HuggingFaceTB/SmolLM2-360M | Head | 2% | 2.08% | 360M | 0.3396% | 1.22M | 358.78M |
| HuggingFaceTB/SmolLM2-360M | Head | 3% | 2.92% | 360M | 0.4755% | 1.71M | 358.29M |
| HuggingFaceTB/SmolLM2-360M | Head | 5% | 5.00% | 360M | 0.8151% | 2.93M | 357.07M |
| HuggingFaceTB/SmolLM2-360M | Head | 10% | 10.00% | 360M | 1.6302% | 5.87M | 354.13M |
| ViT-B/16 | attention heads | 0.10 | 14/144 | ~86.6M | 3.1827% | ~2.76M | ~83.84M |
| DeiT-Tiny | attention heads | 0.10 | 4/36 (11.11%) | ~5.7M | 3.4522% | ~0.20M | ~5.50M |
| Swin-Tiny | attention heads | 0.10 | 14/138 | ~28.3M | 0.9604%-3.1844% | ~0.27M-0.90M | ~27.40M-28.03M |

## 5. Results

### 5.1 WikiText-2 10% attention-head pruning

The main pattern is that MAB methods often keep perplexity degradation low, while several static saliency methods show substantial degradation on particular model families.

*Table 4. WikiText-2, 10% attention-head pruning. Entries are percent perplexity change; lower is better. Note. Static saliency baselines in this table are reported under a fixed calibration/evaluation configuration, while stochastic methods are reported as five-seed means over seeds 1, 2, 3, 42, and 123.*

| Model | Random | Magnitude | Wanda-style | Taylor | Fisher | Budgeted greedy | MAB-UCB | MAB-TS |
|---|---|---|---|---|---|---|---|---|
| gpt2 | 158.91 ± 203.31% | 861.97% | 33.91% | 13.35% | 14.52% | 4.69 ± 2.25% | 1.32 ± 0.67% | 2.08 ± 0.57% |
| gpt2-medium | 13.70 ± 6.71% | 94.97% | 120.58% | 8.38% | 8.78% | 7.59 ± 3.75% | 1.70 ± 1.38% | 0.92 ± 1.53% |
| facebook/opt-125m | 21.43 ± 32.05% | 7.52% | 10.08% | 6.05% | 5.78% | 1.89 ± 1.31% | 0.08 ± 0.26% | 0.38 ± 0.52% |
| facebook/opt-350m | 11.51 ± 5.27% | 57.98% | 36.75% | 12.60% | 7.96% | 0.03 ± 0.67% | -1.37 ± 0.96% | -1.46 ± 1.76% |
| EleutherAI/pythia-160m | 259.03 ± 163.52% | 106.44% | 1080.75% | 23.27% | 70.95% | 13.46 ± 6.62% | 8.62 ± 2.70% | 7.76 ± 1.94% |
| EleutherAI/pythia-410m | 538.26 ± 232.64% | 889.65% | 2932.49% | 209.50% | 197.96% | 222.09 ± 186.35% | 26.58 ± 12.36% | 16.50 ± 2.38% |
| Qwen/Qwen2.5-0.5B | 66.08 ± 53.19% | 35.36% | 21298.44% | 302.34% | 356.85% | 47.75 ± 29.59% | 6.76 ± 2.85% | 11.20 ± 9.44% |

*Table 5. Best MAB method compared with the best non-MAB baseline on WikiText-2 head pruning. Note. Differences are computed from unrounded values and displayed after rounding.*

| Model | Best MAB | Best non-MAB baseline | MAB advantage over best non-MAB |
|---|---|---|---|
| gpt2 | MAB-UCB: 1.32 ± 0.67% | Budgeted greedy: 4.69 ± 2.25% | 3.38 percentage points |
| gpt2-medium | MAB-TS: 0.92 ± 1.53% | Budgeted greedy: 7.59 ± 3.75% | 6.68 percentage points |
| facebook/opt-125m | MAB-UCB: 0.08 ± 0.26% | Budgeted greedy: 1.89 ± 1.31% | 1.80 percentage points |
| facebook/opt-350m | MAB-TS: -1.46 ± 1.76% | Budgeted greedy: 0.03 ± 0.67% | 1.49 percentage points |
| EleutherAI/pythia-160m | MAB-TS: 7.76 ± 1.94% | Budgeted greedy: 13.46 ± 6.62% | 5.69 percentage points |
| EleutherAI/pythia-410m | MAB-TS: 16.50 ± 2.38% | Fisher: 197.96% | 181.46 percentage points |
| Qwen/Qwen2.5-0.5B | MAB-UCB: 6.76 ± 2.85% | Magnitude: 35.36% | 28.60 percentage points |

## 5.2 GPT-2 MLP-group pruning

Table 6 shows GPT-2 MLP-group pruning. This experiment provides an important complementary test because the same pattern holds across multiple pruning ratios. At 3%, 5%, and 8%, MAB-UCB is the best method and produces negative perplexity changes, meaning the selected masks slightly improve evaluation perplexity on the subset. The advantage over budgeted greedy is clear: at 5% MLP pruning, MAB-UCB gives -4.85% compared with -1.99% for greedy; at 8%, MAB-UCB gives -1.65% while greedy gives +2.05%.

*Table 6. WikiText-2, GPT-2 MLP-group pruning. Entries are percent perplexity change; lower is better.*

| GPT-2 MLP prune ratio | Random | Magnitude | Wanda-style | Taylor | Fisher | Budgeted greedy | MAB-UCB | MAB-TS |
|---|---|---|---|---|---|---|---|---|
| 3% | 7.23 ± 3.76% | 8.76% | 3.64% | 4.10% | 5.59% | -4.41 ± 1.02% | -6.25 ± 0.31% | -5.42 ± 1.04% |
| 5% | 11.94 ± 3.74% | 18.92% | 5.62% | 7.62% | 9.45% | -1.99 ± 2.05% | -4.85 ± 0.39% | -4.51 ± 1.78% |
| 8% | 22.35 ± 4.35% | 28.19% | 14.38% | 16.66% | 20.68% | 2.05 ± 2.76% | -1.65 ± 1.19% | -0.99 ± 2.10% |

To complement Table 6, Figure 2 plots GPT-2 MLP-group pruning across ratios. The plot makes the trend clearer: the bandit methods remain below the budgeted greedy curve across the tested ratios, with lower percent perplexity change indicating better preservation.

*Figure 2. GPT-2 MLP-group pruning across pruning ratios on WikiText-2. Points and error bars are computed from the five-seed mean and standard deviation entries in Table 6. Lower percent perplexity change is better.*

## 5.3 LAMBADA cross-dataset evaluation

Table 7 separates LAMBADA from WikiText-2. The masks are selected using WikiText-2 calibration but evaluated on LAMBADA. This tests whether the selected units are only overfit to the calibration/evaluation corpus or whether they transfer to a different context-sensitive language benchmark. MAB gives the lowest degradation on all four models in this table, although the scale of the advantage varies by model. The largest example is Pythia-410M: MAB-TS gives a 30.30% perplexity increase, while budgeted greedy gives 238.85%, random gives 404.96%, and magnitude gives 548.02%. Qwen2.5-0.5B also shows a clear advantage: MAB-UCB gives 10.46%, compared with 30.88% for magnitude and 49.90% for budgeted greedy.

***Table 7. LAMBADA, 10% attention-head pruning, evaluated separately from WikiText-2. Entries are percent perplexity change.***

| Model | Random | Magnitude | Budgeted greedy | MAB-UCB | MAB-TS |
|---|---|---|---|---|---|
| EleutherAI/pythia-410m | 404.96 ± 127.42% | 548.02% | 238.85 ± 207.07% | 38.89 ± 17.19% | 30.30 ± 5.76% |
| Qwen/Qwen2.5-0.5B | 57.33 ± 53.36% | 30.88% | 49.90 ± 36.54% | 10.46 ± 1.16% | 12.66 ± 3.10% |
| facebook/opt-350m | 14.04 ± 9.12% | 75.27% | 7.11 ± 1.81% | 3.51 ± 1.33% | 4.50 ± 1.37% |
| gpt2-medium | 13.60 ± 3.06% | 72.29% | 13.75 ± 7.48% | 7.53 ± 1.96% | 6.70 ± 2.88% |

## 5.4 Matched-budget vision-transformer evaluation on Imagenette

Table 8 is the primary vision comparison because it controls the paired candidate-evaluation budget. For ViT-B/16 and Swin-Tiny, the budgeted greedy baseline and the MAB methods use the same reported number of paired candidate-evaluation trials. Under this matched setting, MAB-TS improves over greedy on ViT-B/16, reducing the Top-1 drop from 3.02 percentage points to 2.05 percentage points. On Swin-Tiny, MAB-TS reduces the Top-1 drop from 2.43 percentage points to 1.34 percentage points, while MAB-UCB gives the lowest reported cross-entropy loss. DeiT-Tiny is retained as a bounded-evaluation case because the available greedy run uses fewer paired trials; it is not used as the main fairness claim.

***Table 8. Matched evaluation-budget analysis on Imagenette. Evals denotes paired candidate-evaluation trials/arm pulls during selection; one paired trial consists of one base and one masked forward evaluation on the same batch. For ViT-B/16 and Swin-Tiny, greedy and MAB use the same number of reported trials. Full details are in Appendix Table A5.***

| Model | Method | Evals | Pruned Top-1 | Δ Top-1 | Pruned loss | Zeroed % |
|---|---|---|---|---|---|---|
| DeiT-Tiny | Budgeted greedy | 138 | 0.7235 ± 0.0260 | -0.0354 ± 0.0237 | 1.1435 ± 0.1194 | 3.4522 |
| DeiT-Tiny | MAB-TS | 256 | 0.7354 ± 0.0104 | -0.0235 ± 0.0110 | 1.0781 ± 0.0336 | 3.4522 |
| DeiT-Tiny | MAB-UCB | 256 | 0.7349 ± 0.0078 | -0.0240 ± 0.0087 | 1.0739 ± 0.0227 | 3.4522 |
| Swin-Tiny | Budgeted greedy | 896 | 0.8444 ± 0.0294 | -0.0243 ± 0.0282 | 0.5221 ± 0.1061 | 2.9064 ± 0.4137 |
| Swin-Tiny | MAB-TS | 896 | 0.8553 ± 0.0077 | -0.0134 ± 0.0072 | 0.4790 ± 0.0272 | 3.1410 ± 0.2702 |
| Swin-Tiny | MAB-UCB | 896 | 0.8544 ± 0.0156 | -0.0143 ± 0.0170 | 0.4649 ± 0.0560 | 3.1671 ± 0.3974 |
| ViT-B/16 | Budgeted greedy | 896 | 0.8226 ± 0.0242 | -0.0302 ± 0.0212 | 0.6985 ± 0.0698 | 3.1827 |
| ViT-B/16 | MAB-TS | 896 | 0.8323 ± 0.0133 | -0.0205 ± 0.0079 | 0.6291 ± 0.0455 | 3.1827 |
| ViT-B/16 | MAB-UCB | 896 | 0.8226 ± 0.0167 | -0.0302 ± 0.0131 | 0.6638 ± 0.0782 | 3.1827 |

Figure 3 visualizes the candidate-evaluation comparison from Table 8. For ViT-B/16 and Swin-Tiny, MAB-TS gives the best reported Top-1 retention when greedy and MAB use the same reported evaluation count. MAB-TS also gives the best reported retention for DeiT-Tiny, but its greedy run uses fewer trials; consequently, DeiT-Tiny is descriptive and is not part of the matched-budget fairness claim.

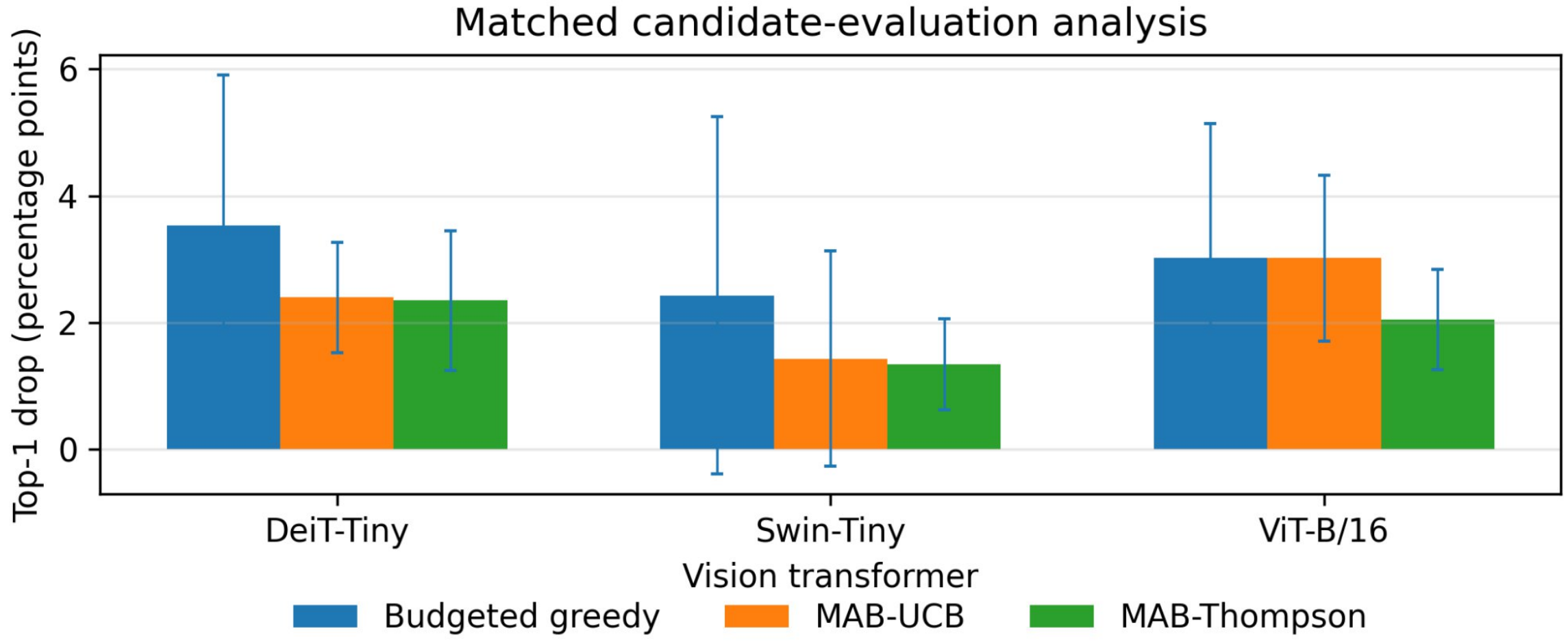

Figure 3. Candidate-evaluation analysis on Imagenette. Evals denote paired candidate-evaluation trials or arm pulls during selection, and lower Top-1 drop is better. Greedy and MAB counts are matched for ViT-B/16 and Swin-Tiny only.

### 5.5 Full vision-transformer screening on Imagenette

Table 9 extends the vision analysis to random and magnitude baselines, a lower-cost budgeted greedy baseline, and the two damage-aware bandit methods. The matched-budget comparison in Table 8 remains the primary fairness comparison for ViT-B/16 and Swin-Tiny. Training images are used for calibration and validation images for evaluation. Entries are five-seed means ± standard deviations under a nominal 10% attention-head ratio; integer rounding gives 4/36 heads (11.11%) for DeiT-Tiny.

***Table 9. Full Imagenette screening comparison under a nominal 10% attention-head ratio for ViT/DeiT/Swin. Integer rounding yields 4/36 heads (11.11%) for DeiT-Tiny. Dense Top-1 is the original unmasked accuracy; pruned values are measured after functional head zeroing. Lower loss and smaller Δ Top-1 are better. Full Top-5, dense-loss, zeroed-parameter, and timing details are in Appendix Table A4.***

| Model | Method | Dense Top-1 | Pruned Top-1 | Δ Top-1 | Pruned loss | Evals |
|---|---|---|---|---|---|---|
| DeiT-Tiny | Budgeted greedy | 0.7589 ± 0.0061 | 0.7153 ± 0.0157 | -0.0436 ± 0.0161 | 1.1660 ± 0.0752 | 128 |
| DeiT-Tiny | MAB-TS | 0.7589 ± 0.0061 | 0.7431 ± 0.0105 | -0.0158 ± 0.0108 | 1.0378 ± 0.0343 | 256 |
| DeiT-Tiny | MAB-UCB | 0.7589 ± 0.0061 | 0.7411 ± 0.0069 | -0.0178 ± 0.0077 | 1.0470 ± 0.0542 | 256 |
| DeiT-Tiny | Magnitude | 0.7589 ± 0.0061 | 0.6650 ± 0.0061 | -0.0939 ± 0.0056 | 1.5104 ± 0.0269 | 0 |
| DeiT-Tiny | Random | 0.7589 ± 0.0061 | 0.7001 ± 0.0153 | -0.0588 ± 0.0212 | 1.3036 ± 0.1418 | 0 |
| Swin-Tiny | Budgeted greedy | 0.8687 ± 0.0026 | 0.8465 ± 0.0148 | -0.0222 ± 0.0133 | 0.5120 ± 0.0533 | 448 |
| Swin-Tiny | MAB-TS | 0.8687 ± 0.0026 | 0.8675 ± 0.0040 | -0.0012 ± 0.0029 | 0.4336 ± 0.0204 | 896 |
| Swin-Tiny | MAB-UCB | 0.8687 ± 0.0026 | 0.8568 ± 0.0101 | -0.0119 ± 0.0094 | 0.4667 ± 0.0220 | 896 |
| Swin-Tiny | Magnitude | 0.8687 ± 0.0026 | 0.1694 ± 0.0032 | -0.6993 ± 0.0046 | 5.3692 ± 0.0395 | 0 |
| Swin-Tiny | Random | 0.8687 ± 0.0026 | 0.8443 ± 0.0103 | -0.0244 ± 0.0104 | 0.6016 ± 0.0829 | 0 |
| ViT-B/16 | Budgeted greedy | 0.8528 ± 0.0067 | 0.8053 ± 0.0234 | -0.0475 ± 0.0199 | 0.7403 ± 0.1017 | 448 |
| ViT-B/16 | MAB-TS | 0.8528 ± 0.0067 | 0.8367 ± 0.0101 | -0.0161 ± 0.0071 | 0.6055 ± 0.0397 | 896 |
| ViT-B/16 | MAB-UCB | 0.8528 ± 0.0067 | 0.8428 ± 0.0062 | -0.0100 ± 0.0094 | 0.5758 ± 0.0265 | 896 |
| ViT-B/16 | Magnitude | 0.8528 ± 0.0067 | 0.8177 ± 0.0041 | -0.0351 ± 0.0038 | 0.7375 ± 0.0163 | 0 |
| ViT-B/16 | Random | 0.8528 ± 0.0067 | 0.8053 ± 0.0196 | -0.0475 ± 0.0172 | 0.7487 ± 0.0937 | 0 |

Figure 4 visualizes the adaptive methods from the full vision-transformer screening experiment. Magnitude is omitted because its severe collapse on Swin-Tiny dominates the scale; the complete all-method numerical record is provided in Appendix Table A4.

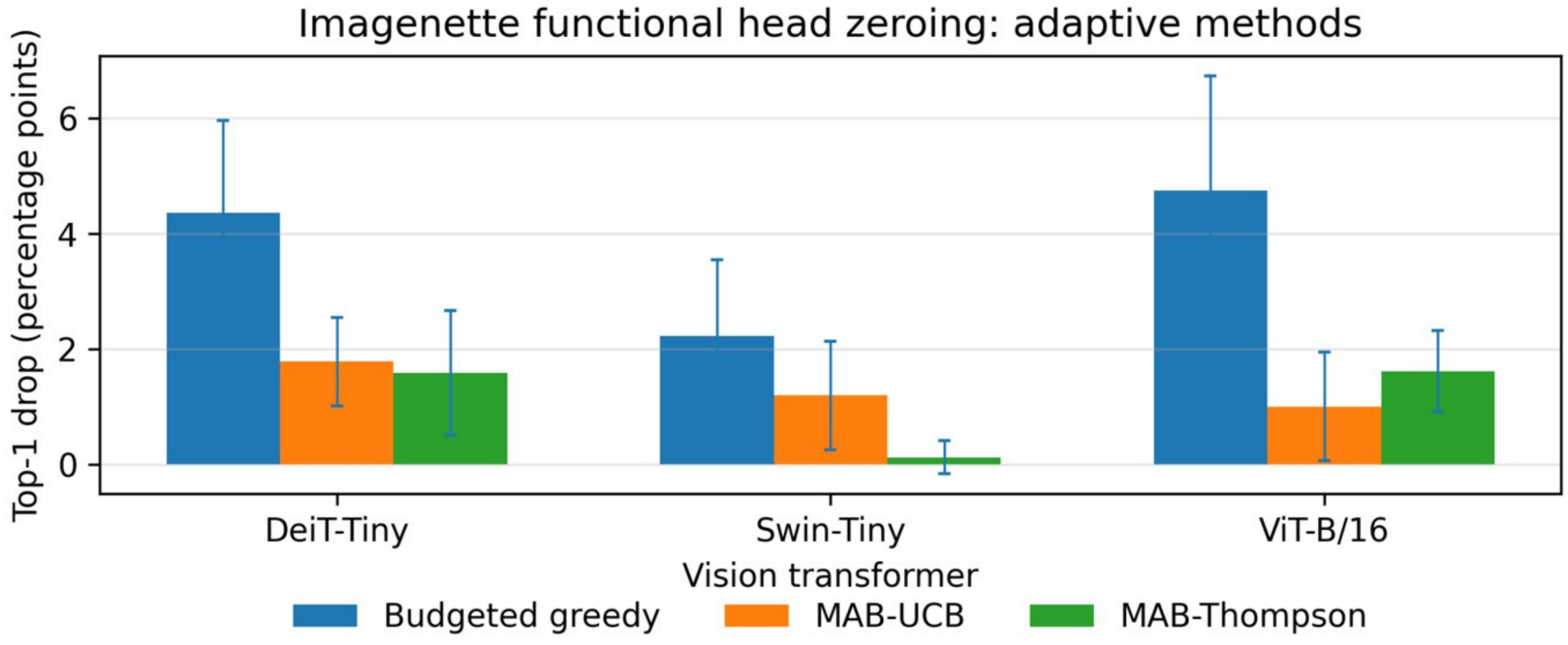

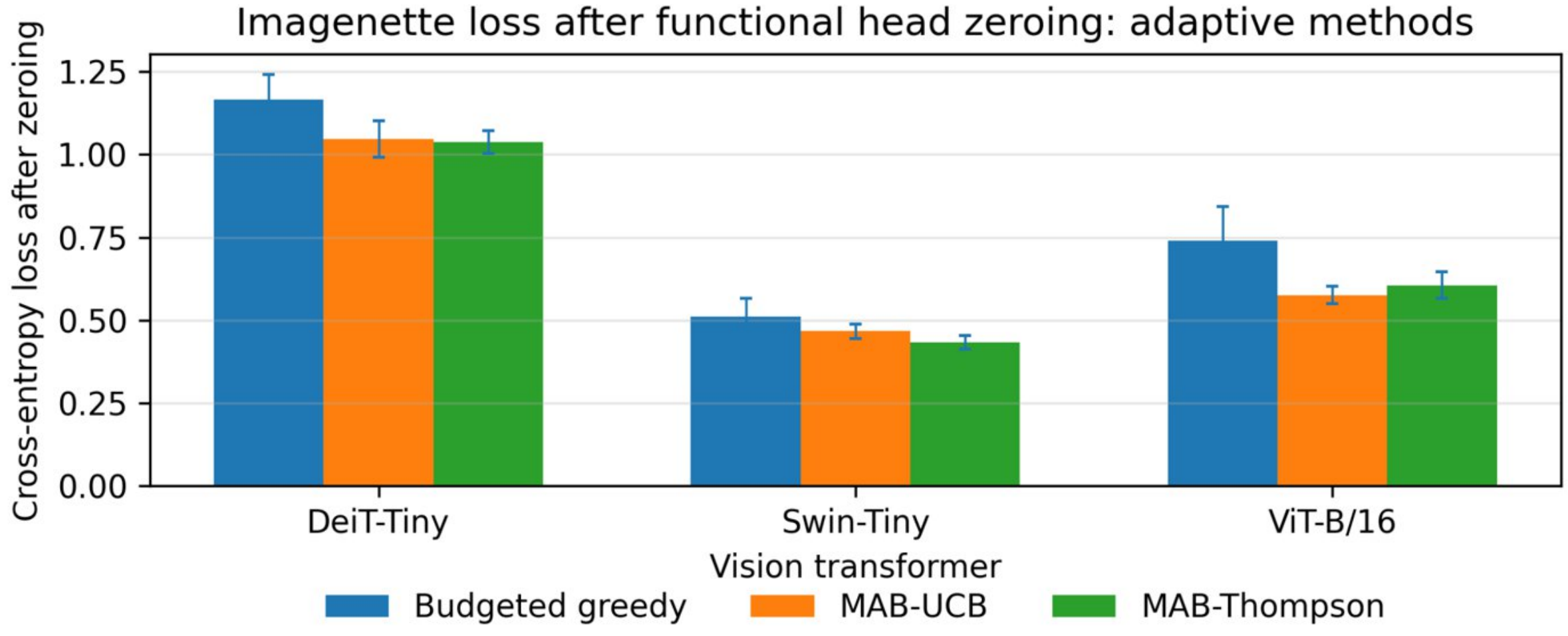

Figure 4. Imagenette functional attention-head zeroing with adaptive methods. Top: Top-1 accuracy drop in percentage points. Bottom: cross-entropy loss after zeroing. Error bars show variation across the five seeds used in the vision experiments.

## 5.6 SmolLM2 stability and NaN handling

SmolLM2-360M is reported as a sensitivity case rather than a primary benchmark table.

*Table 10. SmolLM2-360M sensitivity analysis. Non-finite perplexity runs are treated as failures and excluded from the finite mean; valid runs are reported explicitly.*

| Ratio | Actual unit ratio | Effective zeroed params | Method | PPL change (finite mean ± std) | Valid runs |
|---|---|---|---|---|---|
| 2% | 2.08% | 0.3396% | Random | 13.44 ± 16.40% | 5/5 |
| 2% | 2.08% | 0.3396% | Budgeted greedy | 24.06 ± 18.08% | 3/5 |

| Ratio | Actual unit ratio | Effective zeroed params | Method | PPL change (finite mean ± std) | Valid runs |
|---|---|---|---|---|---|
| 2% | 2.08% | 0.3396% | MAB-UCB | 1.68 ± 0.85% | 5/5 |
| 2% | 2.08% | 0.3396% | MAB-TS | 1.29 ± 0.49% | 4/5 |
| 3% | 2.92% | 0.4755% | Random | 20.84 ± 20.43% | 4/5 |
| 3% | 2.92% | 0.4755% | Budgeted greedy | 27.07 ± 20.91% | 3/5 |
| 3% | 2.92% | 0.4755% | MAB-UCB | 2.40 ± 0.95% | 5/5 |
| 3% | 2.92% | 0.4755% | MAB-TS | 1.81 ± 0.61% | 4/5 |
| 5% | 5.00% | 0.8151% | Random | 45.56 ± 58.77% | 4/5 |
| 5% | 5.00% | 0.8151% | Magnitude | NaN/failure | 0/3 |
| 5% | 5.00% | 0.8151% | Budgeted greedy | 57.40 ± 59.97% | 3/5 |
| 5% | 5.00% | 0.8151% | MAB-UCB | 12.67 ± 21.13% | 5/5 |
| 5% | 5.00% | 0.8151% | MAB-TS | 3.46 ± 0.81% | 4/5 |
| 10% | 10.00% | 1.6302% | Random | NaN/failure | 0/3 |
| 10% | 10.00% | 1.6302% | Magnitude | NaN/failure | 0/3 |
| 10% | 10.00% | 1.6302% | Budgeted greedy | NaN/failure | 0/3 |
| 10% | 10.00% | 1.6302% | MAB-UCB | 99.07 ± 146.99% | 3/3 |
| 10% | 10.00% | 1.6302% | MAB-TS | 67.19 ± 100.84% | 3/3 |
| 10% | 10.00% | 1.6302% | Taylor | NaN/failure | 0/3 |
| 10% | 10.00% | 1.6302% | Fisher | NaN/failure | 0/3 |
| 10% | 10.00% | 1.6302% | Wanda-style | NaN/failure | 0/3 |

### 5.7 Statistical analysis and reproducibility

To avoid relying only on visual trends, the language-model results were analysed using dataset-wise paired comparisons. For each dataset, model, target, and pruning ratio, MAB methods were compared with budgeted greedy using seed-level paired differences in percent perplexity change, bootstrap 95% confidence intervals, paired tests, and the Benjamini-Hochberg false-discovery-rate procedure [29]. The correction was applied across the full family of 116 dataset-wise tests. Among the 28 MAB-versus-greedy comparisons highlighted in the paper, 23 bootstrap confidence intervals exclude zero, 11 paired tests have $p < 0.05$, and 6 have $q < 0.05$ under that 116-test correction. These counts should be interpreted conservatively and do not imply that every comparison is significant.

*Table 11. Summary of the 28 highlighted paired language-model comparisons against budgeted greedy. The q-values use Benjamini-Hochberg correction across the full family of 116 dataset-wise tests.*

| Metric | Value |
|---|---|
| Highlighted MAB vs greedy rows | 28 |
| Bootstrap 95% CI excludes zero | 23/28 |
| Paired $p < 0.05$ | 11/28 |
| BH $q < 0.05$ (116-test family) | 6/28 |

Result summaries are stratified by dataset, model, target, ratio, method, and seed. The Evals column counts adaptive paired candidate trials, not activation or gradient passes used to construct static saliency scores. Thus, Wanda-style, Taylor, and Fisher may use calibration computations while retaining Evals = 0; this convention is applied throughout the main and appendix tables.

## 6. Discussion

Across the reported experiments, the bandit methods frequently reduce degradation relative to budgeted greedy, but the evidence is not uniform across all configurations. In the 28 highlighted language comparisons, 11 paired tests have $p < 0.05$ and 6 remain below $q < 0.05$ after correction across 116 tests. The results therefore support improved selection in several configurations rather than a universal advantage. The non-finite SmolLM2 runs further show that stability depends on the model and pruning setting.

The matched-evaluation vision comparison helps separate the allocation rule from the number of paired trials. For ViT-B/16 and Swin-Tiny, greedy and MAB use the same reported trial count, and a bandit method yields the lower reported degradation. DeiT-Tiny is not included in that fairness claim because its greedy run uses fewer evaluations.

Together with the language results, this indicates that adaptive allocation can be useful under a bounded search budget, although the limited benchmark scale and five-seed design constrain generalisation.

Relative to existing structured transformer pruning [10-14], the present study focuses on evaluation-budget allocation and paired damage rather than physical architecture reduction. Relative to the authors' earlier MAB pruning work [15-17], it introduces typed transformer units and sequential conditioning on the cumulative mask. Its deployment value remains to be established through physical model surgery, broader experimental baselines, larger benchmarks, and hardware measurements.

## 7. Limitations

- The experiments use functional zeroing/masking. Dense checkpoint tensors remain stored, so the effective zeroed-parameter counts do not represent physical model-size or memory reduction, measured deployment speedup, or hardware acceleration. Obtaining those benefits requires separate model surgery and hardware evaluation.
- The vision-transformer experiments use Imagenette rather than full ImageNet-1K. They provide ViT/DeiT/Swin evidence and a matched-evaluation analysis for ViT-B/16 and Swin-Tiny, but future work should repeat the protocol on ImageNet-1K or another large-scale visual-recognition benchmark.
- SmolLM2 shows non-finite perplexity in some configurations; these are reported as stability failures and not averaged away.
- The bandit reward estimates marginal single-unit damage. Interactions among multiple units in the final top-k mask are not explicitly optimized during search and are assessed only by the final empirical evaluation.
- The method evaluates post-training pruning only. It does not include fine-tuning or recovery training after pruning.
- The main language search is restricted to a low-magnitude pre-screened pool of at most 48 candidates. The present experiments do not isolate the contribution of this screening step from the subsequent bandit allocation.
- The study does not report ablations over reward temperature, active-pool size, per-step pull budget, or MLP group width, so robustness to these choices remains unquantified.
- The experimental baselines do not include direct implementations of recent structured transformer-pruning methods [10-14]. These methods differ in objective, recovery training, and physical compression, so a harmonised comparison remains future work.

## 8. Conclusion

This paper formulated post-training transformer structured-unit selection as a fixed-budget damage-aware bandit problem. Attention heads and MLP groups are evaluated by paired same-batch loss changes, and MAB-UCB or MAB-TS constructs the final mask sequentially. Across the reported language and vision experiments, bandit selection reduces degradation in several configurations, including matched-evaluation results for ViT-B/16 and Swin-Tiny; however, only a subset of the highlighted paired comparisons remains significant after multiple-testing correction. Because the implementation uses functional zeroing, the findings concern selection quality rather than realised compression or speedup. Future work should implement physical model surgery, evaluate latency and memory, compare directly with recent structured-pruning methods, and extend the study to ImageNet-1K and larger language models.

## Code Availability

The source code used in this study is publicly available at: https://github.com/SalemAmeen/mab-structured-pruning

The repository contains the language-model and vision-transformer pruning implementations, experiment-running scripts, selected result summaries and figures, dependency files, and reproducibility documentation. It supports functional structured zeroing for transformer attention heads and MLP groups, together with random, magnitude, budgeted greedy, MAB-UCB, and MAB-TS selection. The repository is released under the MIT License.

## Declaration of generative AI and AI-assisted technologies in the manuscript preparation process

During the preparation of this work, the authors used ChatGPT (OpenAI) to assist with language editing, restructuring, rephrasing, and literature organisation. After using this tool, the authors reviewed and edited the content as needed and take full responsibility for the content of the published article.

## Appendix A. Complete detailed result tables

Appendix Tables A1-A5 provide the detailed per-method records underlying the summary tables, including final perplexity or accuracy, percentage change, pruning target and ratio, effective zeroed-parameter estimates, and evaluation counts.

***Appendix Table A1. Detailed WikiText-2 10% head-pruning results.***

| Dataset | Model | Target | Ratio | Method | Final PPL | PPL change | Effective zeroed params | Evals/time note |
|---|---|---|---|---|---|---|---|---|
| WikiText-2 | gpt2 | Head | 10% | Random | 164.4357 | 158.91 ± 203.31% | 2.2141% | Static/no selection evals |
| WikiText-2 | gpt2 | Head | 10% | Magnitude | 610.9633 | 861.97% | 2.2141% | Static/no selection evals |
| WikiText-2 | gpt2 | Head | 10% | Wanda-style | 84.4646 | 33.91% | 2.2141% | Static/no selection evals |
| WikiText-2 | gpt2 | Head | 10% | Taylor | 71.4956 | 13.35% | 2.2141% | Static/no selection evals |
| WikiText-2 | gpt2 | Head | 10% | Fisher | 72.2332 | 14.52% | 2.2141% | Static/no selection evals |
| WikiText-2 | gpt2 | Head | 10% | Budgeted greedy | 66.4915 | 4.69 ± 2.25% | 2.2141% | MAB/greedy selection |
| WikiText-2 | gpt2 | Head | 10% | MAB-UCB | 64.3470 | 1.32 ± 0.67% | 2.2141% | MAB/greedy selection |
| WikiText-2 | gpt2 | Head | 10% | MAB-TS | 64.8309 | 2.08 ± 0.57% | 2.2141% | MAB/greedy selection |
| WikiText-2 | gpt2-medium | Head | 10% | Random | 51.1782 | 13.70 ± 6.71% | 2.8095% | Static/no selection evals |
| WikiText-2 | gpt2-medium | Head | 10% | Magnitude | 87.7627 | 94.97% | 2.8095% | Static/no selection evals |
| WikiText-2 | gpt2-medium | Head | 10% | Wanda-style | 99.2900 | 120.58% | 2.8095% | Static/no selection evals |
| WikiText-2 | gpt2-medium | Head | 10% | Taylor | 48.7872 | 8.38% | 2.8095% | Static/no selection evals |
| WikiText-2 | gpt2-medium | Head | 10% | Fisher | 48.9653 | 8.78% | 2.8095% | Static/no selection evals |
| WikiText-2 | gpt2-medium | Head | 10% | Budgeted greedy | 48.4308 | 7.59 ± 3.75% | 2.8095% | MAB/greedy selection |
| WikiText-2 | gpt2-medium | Head | 10% | MAB-UCB | 45.7761 | 1.70 ± 1.38% | 2.8095% | MAB/greedy selection |
| WikiText-2 | gpt2-medium | Head | 10% | MAB-TS | 45.4260 | 0.92 ± 1.53% | 2.8095% | MAB/greedy selection |
| WikiText-2 | facebook/opt-125m | Head | 10% | Random | 91.7061 | 21.43 ± 32.05% | 2.1999% | Static/no selection evals |
| WikiText-2 | facebook/opt-125m | Head | 10% | Magnitude | 81.1991 | 7.52% | 2.1999% | Static/no selection evals |
| WikiText-2 | facebook/opt-125m | Head | 10% | Wanda-style | 81.7277 | 10.08% | 2.1999% | Static/no selection evals |
| WikiText-2 | facebook/opt-125m | Head | 10% | Taylor | 78.7354 | 6.05% | 2.1999% | Static/no selection evals |
| WikiText-2 | facebook/opt-125m | Head | 10% | Fisher | 78.5348 | 5.78% | 2.1999% | Static/no selection evals |
| WikiText-2 | facebook/opt-125m | Head | 10% | Budgeted greedy | 76.9468 | 1.89 ± 1.31% | 2.1999% | MAB/greedy selection |
| WikiText-2 | facebook/opt-125m | Head | 10% | MAB-UCB | 75.5854 | 0.08 ± 0.26% | 2.1999% | MAB/greedy selection |
| WikiText-2 | facebook/opt-125m | Head | 10% | MAB-TS | 75.8114 | 0.38 ± 0.52% | 2.1999% | MAB/greedy selection |
| WikiText-2 | facebook/opt-350m | Head | 10% | Random | 65.6206 | 11.51 ± 5.27% | 3.0099% | Static/no selection evals |
| WikiText-2 | facebook/opt-350m | Head | 10% | Magnitude | 92.9654 | 57.98% | 3.0099% | Static/no selection evals |
| WikiText-2 | facebook/opt-350m | Head | 10% | Wanda-style | 80.4697 | 36.75% | 3.0099% | Static/no selection evals |
| WikiText-2 | facebook/opt-350m | Head | 10% | Taylor | 66.2626 | 12.60% | 3.0099% | Static/no selection evals |
| WikiText-2 | facebook/opt-350m | Head | 10% | Fisher | 63.5288 | 7.96% | 3.0099% | Static/no selection evals |
| WikiText-2 | facebook/opt-350m | Head | 10% | Budgeted greedy | 58.8644 | 0.03 ± 0.67% | 3.0099% | MAB/greedy selection |
| WikiText-2 | facebook/opt-350m | Head | 10% | MAB-UCB | 58.0382 | -1.37 ± 0.96% | 3.0099% | MAB/greedy selection |
| WikiText-2 | facebook/opt-350m | Head | 10% | MAB-TS | 57.9865 | -1.46 ± 1.76% | 3.0099% | MAB/greedy selection |
| WikiText-2 | EleutherAI/pythia-160m | Head | 10% | Random | 247.7770 | 259.03 ± 163.52% | 1.6974% | Static/no selection evals |

| Dataset | Model | Target | Ratio | Method | Final PPL | PPL change | Effective zeroed params | Evals/time note |
|---|---|---|---|---|---|---|---|---|
| WikiText-2 | EleutherAI/pythia-160m | Head | 10% | Magnitude | 142.4696 | 106.44% | 1.6974% | Static/no selection evals |
| WikiText-2 | EleutherAI/pythia-160m | Head | 10% | Wanda-style | 815.0906 | 1080.75% | 1.6974% | Static/no selection evals |
| WikiText-2 | EleutherAI/pythia-160m | Head | 10% | Taylor | 85.0937 | 23.27% | 1.6974% | Static/no selection evals |
| WikiText-2 | EleutherAI/pythia-160m | Head | 10% | Fisher | 118.0075 | 70.95% | 1.6974% | Static/no selection evals |
| WikiText-2 | EleutherAI/pythia-160m | Head | 10% | Budgeted greedy | 78.3010 | 13.46 ± 6.62% | 1.6974% | MAB/greedy selection |
| WikiText-2 | EleutherAI/pythia-160m | Head | 10% | MAB-UCB | 74.9603 | 8.62 ± 2.70% | 1.6974% | MAB/greedy selection |
| WikiText-2 | EleutherAI/pythia-160m | Head | 10% | MAB-TS | 74.3707 | 7.76 ± 1.94% | 1.6974% | MAB/greedy selection |
| WikiText-2 | EleutherAI/pythia-410m | Head | 10% | Random | 257.8930 | 538.26 ± 232.64% | 2.4594% | Static/no selection evals |
| WikiText-2 | EleutherAI/pythia-410m | Head | 10% | Magnitude | 399.8753 | 889.65% | 2.4594% | Static/no selection evals |
| WikiText-2 | EleutherAI/pythia-410m | Head | 10% | Wanda-style | 1225.2974 | 2932.49% | 2.4594% | Static/no selection evals |
| WikiText-2 | EleutherAI/pythia-410m | Head | 10% | Taylor | 125.0559 | 209.50% | 2.4594% | Static/no selection evals |
| WikiText-2 | EleutherAI/pythia-410m | Head | 10% | Fisher | 120.3946 | 197.96% | 2.4594% | Static/no selection evals |
| WikiText-2 | EleutherAI/pythia-410m | Head | 10% | Budgeted greedy | 130.1414 | 222.09 ± 186.35% | 2.4594% | MAB/greedy selection |
| WikiText-2 | EleutherAI/pythia-410m | Head | 10% | MAB-UCB | 51.1461 | 26.58 ± 12.36% | 2.4594% | MAB/greedy selection |
| WikiText-2 | EleutherAI/pythia-410m | Head | 10% | MAB-TS | 47.0744 | 16.50 ± 2.38% | 2.4594% | MAB/greedy selection |
| WikiText-2 | Qwen/Qwen2.5-0.5B | Head | 10% | Random | 50.3481 | 66.08 ± 53.19% | 0.7893% | Static/no selection evals |
| WikiText-2 | Qwen/Qwen2.5-0.5B | Head | 10% | Magnitude | 41.0363 | 35.36% | 0.7893% | Static/no selection evals |
| WikiText-2 | Qwen/Qwen2.5-0.5B | Head | 10% | Wanda-style | 6487.1669 | 21298.44% | 0.7893% | Static/no selection evals |
| WikiText-2 | Qwen/Qwen2.5-0.5B | Head | 10% | Taylor | 121.9728 | 302.34% | 0.7893% | Static/no selection evals |
| WikiText-2 | Qwen/Qwen2.5-0.5B | Head | 10% | Fisher | 138.4980 | 356.85% | 0.7893% | Static/no selection evals |
| WikiText-2 | Qwen/Qwen2.5-0.5B | Head | 10% | Budgeted greedy | 44.7921 | 47.75 ± 29.59% | 0.7893% | MAB/greedy selection |
| WikiText-2 | Qwen/Qwen2.5-0.5B | Head | 10% | MAB-UCB | 32.3651 | 6.76 ± 2.85% | 0.7893% | MAB/greedy selection |
| WikiText-2 | Qwen/Qwen2.5-0.5B | Head | 10% | MAB-TS | 33.7120 | 11.20 ± 9.44% | 0.7893% | MAB/greedy selection |

### *Appendix Table A2. Detailed WikiText-2 GPT-2 MLP-group pruning results.*

| Dataset | Model | Target | Ratio | Method | Final PPL | PPL change | Effective zeroed params |
|---|---|---|---|---|---|---|---|
| WikiText-2 | gpt2 | MLP group | 3% | Random | 68.1063 | 7.23 ± 3.76% | 1.3834% |
| WikiText-2 | gpt2 | MLP group | 3% | Magnitude | 69.0764 | 8.76% | 1.3834% |
| WikiText-2 | gpt2 | MLP group | 3% | Wanda-style | 65.8220 | 3.64% | 1.3834% |
| WikiText-2 | gpt2 | MLP group | 3% | Taylor | 66.1139 | 4.10% | 1.3834% |
| WikiText-2 | gpt2 | MLP group | 3% | Fisher | 67.0646 | 5.59% | 1.3834% |
| WikiText-2 | gpt2 | MLP group | 3% | Budgeted greedy | 60.7080 | -4.41 ± 1.02% | 1.3834% |
| WikiText-2 | gpt2 | MLP group | 3% | MAB-UCB | 59.5413 | -6.25 ± 0.31% | 1.3834% |
| WikiText-2 | gpt2 | MLP group | 3% | MAB-TS | 60.0661 | -5.42 ± 1.04% | 1.3834% |
| WikiText-2 | gpt2 | MLP group | 5% | Random | 71.0917 | 11.94 ± 3.74% | 2.2924% |
| WikiText-2 | gpt2 | MLP group | 5% | Magnitude | 75.5275 | 18.92% | 2.2924% |
| WikiText-2 | gpt2 | MLP group | 5% | Wanda-style | 67.0797 | 5.62% | 2.2924% |
| WikiText-2 | gpt2 | MLP group | 5% | Taylor | 68.3541 | 7.62% | 2.2924% |
| WikiText-2 | gpt2 | MLP group | 5% | Fisher | 69.5143 | 9.45% | 2.2924% |
| WikiText-2 | gpt2 | MLP group | 5% | Budgeted greedy | 62.2458 | -1.99 ± 2.05% | 2.2924% |
| WikiText-2 | gpt2 | MLP group | 5% | MAB-UCB | 60.4282 | -4.85 ± 0.39% | 2.2924% |
| WikiText-2 | gpt2 | MLP group | 5% | MAB-TS | 60.6444 | -4.51 ± 1.78% | 2.2924% |
| WikiText-2 | gpt2 | MLP group | 8% | Random | 77.7046 | 22.35 ± 4.35% | 3.6362% |
| WikiText-2 | gpt2 | MLP group | 8% | Magnitude | 81.4165 | 28.19% | 3.6362% |
| WikiText-2 | gpt2 | MLP group | 8% | Wanda-style | 72.6435 | 14.38% | 3.6362% |
| WikiText-2 | gpt2 | MLP group | 8% | Taylor | 74.0912 | 16.66% | 3.6362% |
| WikiText-2 | gpt2 | MLP group | 8% | Fisher | 76.6459 | 20.68% | 3.6362% |
| WikiText-2 | gpt2 | MLP group | 8% | Budgeted greedy | 64.8143 | 2.05 ± 2.76% | 3.6362% |

| Dataset | Model | Target | Ratio | Method | Final PPL | PPL change | Effective zeroed params |
|---|---|---|---|---|---|---|---|
| WikiText-2 | gpt2 | MLP group | 8% | MAB-UCB | 62.4622 | -1.65 ± 1.19% | 3.6362% |
| WikiText-2 | gpt2 | MLP group | 8% | MAB-TS | 62.8817 | -0.99 ± 2.10% | 3.6362% |

***Appendix Table A3. Detailed LAMBADA 10% head-pruning results.***

| Dataset | Model | Target | Ratio | Method | Final PPL | PPL change | Effective zeroed params |
|---|---|---|---|---|---|---|---|
| LAMBADA | EleutherAI/pythia-410m | Head | 10% | Random | 241.1135 | 404.96 ± 127.42% | 2.4594% |
| LAMBADA | EleutherAI/pythia-410m | Head | 10% | Magnitude | 309.4229 | 548.02% | 2.4594% |
| LAMBADA | EleutherAI/pythia-410m | Head | 10% | Budgeted greedy | 161.8000 | 238.85 ± 207.07% | 2.4594% |
| LAMBADA | EleutherAI/pythia-410m | Head | 10% | MAB-UCB | 66.3192 | 38.89 ± 17.19% | 2.4594% |
| LAMBADA | EleutherAI/pythia-410m | Head | 10% | MAB-TS | 62.2185 | 30.30 ± 5.76% | 2.4594% |
| LAMBADA | Qwen/Qwen2.5-0.5B | Head | 10% | Random | 63.1751 | 57.33 ± 53.36% | 0.7893% |
| LAMBADA | Qwen/Qwen2.5-0.5B | Head | 10% | Magnitude | 52.5542 | 30.88% | 0.7893% |
| LAMBADA | Qwen/Qwen2.5-0.5B | Head | 10% | Budgeted greedy | 60.1914 | 49.90 ± 36.54% | 0.7893% |
| LAMBADA | Qwen/Qwen2.5-0.5B | Head | 10% | MAB-UCB | 44.3554 | 10.46 ± 1.16% | 0.7893% |
| LAMBADA | Qwen/Qwen2.5-0.5B | Head | 10% | MAB-TS | 45.2381 | 12.66 ± 3.10% | 0.7893% |
| LAMBADA | facebook/opt-350m | Head | 10% | Random | 63.1287 | 14.04 ± 9.12% | 3.0099% |
| LAMBADA | facebook/opt-350m | Head | 10% | Magnitude | 97.0235 | 75.27% | 3.0099% |
| LAMBADA | facebook/opt-350m | Head | 10% | Budgeted greedy | 59.2906 | 7.11 ± 1.81% | 3.0099% |
| LAMBADA | facebook/opt-350m | Head | 10% | MAB-UCB | 57.2988 | 3.51 ± 1.33% | 3.0099% |
| LAMBADA | facebook/opt-350m | Head | 10% | MAB-TS | 57.8471 | 4.50 ± 1.37% | 3.0099% |
| LAMBADA | gpt2-medium | Head | 10% | Random | 60.7929 | 13.60 ± 3.06% | 2.8095% |
| LAMBADA | gpt2-medium | Head | 10% | Magnitude | 92.2028 | 72.29% | 2.8095% |
| LAMBADA | gpt2-medium | Head | 10% | Budgeted greedy | 60.8716 | 13.75 ± 7.48% | 2.8095% |
| LAMBADA | gpt2-medium | Head | 10% | MAB-UCB | 57.5456 | 7.53 ± 1.96% | 2.8095% |
| LAMBADA | gpt2-medium | Head | 10% | MAB-TS | 57.1003 | 6.70 ± 2.88% | 2.8095% |

***Appendix Table A4. Detailed Imagenette ViT/DeiT/Swin attention-head pruning results. The ratio column is nominal; for DeiT-Tiny, 0.10 corresponds to 4/36 heads (11.11%). Evals denotes paired candidate-evaluation trials or arm pulls, each using one base and one masked forward evaluation on the same batch.***

| Dataset | Model | Target | Ratio | Method | Loss after | Δ loss | Top-1 after | Δ Top-1 | Top-5 after | Δ Top-5 | Zeroed % | Evals | Time (s) |
|---|---|---|---|---|---|---|---|---|---|---|---|---|---|
| Imagenette | deit_tiny_patch16_224 | attention heads | 0.10 | Budgeted greedy | 1.1660 ± 0.0752 | 0.1624 ± 0.0748 | 0.7153 ± 0.0157 | -0.0436 ± 0.0161 | 0.9190 ± 0.0140 | -0.0240 ± 0.0126 | 3.4522 | 128 | 3.28 ± 0.06 |
| Imagenette | deit_tiny_patch16_224 | attention heads | 0.10 | MAB-TS | 1.0378 ± 0.0343 | 0.0343 ± 0.0424 | 0.7431 ± 0.0105 | -0.0158 ± 0.0108 | 0.9331 ± 0.0068 | -0.0099 ± 0.0066 | 3.4522 | 256 | 6.62 ± 0.08 |
| Imagenette | deit_tiny_patch16_224 | attention heads | 0.10 | MAB-UCB | 1.0470 ± 0.0542 | 0.0435 ± 0.0401 | 0.7411 ± 0.0069 | -0.0178 ± 0.0077 | 0.9312 ± 0.0070 | -0.0118 ± 0.0041 | 3.4522 | 256 | 6.61 ± 0.12 |
| Imagenette | deit_tiny_patch16_224 | attention heads | 0.10 | magnitude | 1.5104 ± 0.0269 | 0.5069 ± 0.0149 | 0.6650 ± 0.0061 | -0.0939 ± 0.0056 | 0.8756 ± 0.0052 | -0.0674 ± 0.0036 | 3.4522 | 0 | 0.01 ± 0.00 |
| Imagenette | deit_tiny_patch16_224 | attention heads | 0.10 | random | 1.3036 ± 0.1418 | 0.3001 ± 0.1642 | 0.7001 ± 0.0153 | -0.0588 ± 0.0212 | 0.9053 ± 0.0203 | -0.0377 ± 0.0230 | 3.4522 | 0 | 0.00 ± 0.00 |
| Imagenette | swin_tiny_patch4_window7_224 | attention heads | 0.10 | Budgeted greedy | 0.5120 ± 0.0533 | 0.0132 ± 0.0422 | 0.8465 ± 0.0148 | -0.0222 ± 0.0133 | 0.9780 ± 0.0066 | -0.0115 ± 0.0064 | 2.8891 ± 0.4082 | 448 | 47.37 ± 0.10 |
| Imagenette | swin_tiny_patch4_window7_224 | attention heads | 0.10 | MAB-TS | 0.4336 ± 0.020 | -0.0652 ± | 0.8675 ± 0.004 | -0.0012 ± | 0.9848 ± 0.002 | -0.0047 ± | 3.1844 ± 0.277 | 896 | 95.24 ± 0.55 |

| Dataset | Model | Target | Ratio | Method | Loss after | Δ loss | Top-1 after | Δ Top-1 | Top-5 after | Δ Top-5 | Zeroed % | Evals | Time (s) |
|---|---|---|---|---|---|---|---|---|---|---|---|---|---|
| | | | | | 4 | 0.0140 | 0 | 0.0029 | 4 | 0.0023 | 1 | | |
| Imagenette | swin_tiny_patch4_window7_224 | attention heads | 0.10 | MAB-UCB | 0.4667 ± 0.0220 | -0.0321 ± 0.0211 | 0.8568 ± 0.0101 | -0.0119 ± 0.0094 | 0.9827 ± 0.0025 | -0.0068 ± 0.0009 | 3.1063 ± 0.2635 | 896 | 93.99 ± 0.22 |
| Imagenette | swin_tiny_patch4_window7_224 | attention heads | 0.10 | magnitude | 5.3692 ± 0.0395 | 4.8704 ± 0.0305 | 0.1694 ± 0.0032 | -0.6993 ± 0.0046 | 0.3034 ± 0.0059 | -0.6861 ± 0.0053 | 0.9604 | 0 | 0.02 ± 0.00 |
| Imagenette | swin_tiny_patch4_window7_224 | attention heads | 0.10 | random | 0.6016 ± 0.0829 | 0.1028 ± 0.0801 | 0.8443 ± 0.0103 | -0.0244 ± 0.0104 | 0.9806 ± 0.0029 | -0.0089 ± 0.0036 | 3.0715 ± 0.3642 | 0 | 0.00 ± 0.00 |
| Imagenette | vit_base_patch16_224 | attention heads | 0.10 | Budgeted greedy | 0.7403 ± 0.1017 | 0.1802 ± 0.0934 | 0.8053 ± 0.0234 | -0.0475 ± 0.0199 | 0.9607 ± 0.0103 | -0.0216 ± 0.0083 | 3.1827 | 448 | 105.81 ± 1.20 |
| Imagenette | vit_base_patch16_224 | attention heads | 0.10 | MAB-TS | 0.6055 ± 0.0397 | 0.0455 ± 0.0327 | 0.8367 ± 0.0101 | -0.0161 ± 0.0071 | 0.9734 ± 0.0044 | -0.0089 ± 0.0039 | 3.1827 | 896 | 214.71 ± 4.67 |
| Imagenette | vit_base_patch16_224 | attention heads | 0.10 | MAB-UCB | 0.5758 ± 0.0265 | 0.0157 ± 0.0338 | 0.8428 ± 0.0062 | -0.0100 ± 0.0094 | 0.9777 ± 0.0035 | -0.0046 ± 0.0042 | 3.1827 | 896 | 211.07 ± 2.31 |
| Imagenette | vit_base_patch16_224 | attention heads | 0.10 | magnitude | 0.7375 ± 0.0163 | 0.1774 ± 0.0096 | 0.8177 ± 0.0041 | -0.0351 ± 0.0038 | 0.9576 ± 0.0019 | -0.0247 ± 0.0014 | 3.1827 | 0 | 0.02 ± 0.01 |
| Imagenette | vit_base_patch16_224 | attention heads | 0.10 | random | 0.7487 ± 0.0937 | 0.1886 ± 0.0882 | 0.8053 ± 0.0196 | -0.0475 ± 0.0172 | 0.9631 ± 0.0081 | -0.0192 ± 0.0064 | 3.1827 | 0 | 0.00 ± 0.00 |

*Appendix Table A5. Detailed matched-evaluation Imagenette analysis. For ViT-B/16 and Swin-Tiny, greedy and MAB methods use the same reported number of paired candidate-evaluation trials; DeiT-Tiny is included as a bounded-evaluation case.*

| Model | Method | Evals | Loss after | Δ loss | Top-1 after | Δ Top-1 | Top-5 after | Zeroed % |
|---|---|---|---|---|---|---|---|---|
| DeiT-Tiny | Budgeted greedy | 138 | 1.1435 ± 0.1194 | 0.1399 ± 0.1083 | 0.7235 ± 0.0260 | -0.0354 ± 0.0237 | 0.9205 ± 0.0135 | 3.4522 |
| DeiT-Tiny | MAB-TS | 256 | 1.0781 ± 0.0336 | 0.0746 ± 0.0255 | 0.7354 ± 0.0104 | -0.0235 ± 0.0110 | 0.9272 ± 0.0071 | 3.4522 |
| DeiT-Tiny | MAB-UCB | 256 | 1.0739 ± 0.0227 | 0.0703 ± 0.0249 | 0.7349 ± 0.0078 | -0.0240 ± 0.0087 | 0.9304 ± 0.0023 | 3.4522 |
| Swin-Tiny | Budgeted greedy | 896 | 0.5221 ± 0.1061 | 0.0234 ± 0.1019 | 0.8444 ± 0.0294 | -0.0243 ± 0.0282 | 0.9762 ± 0.0091 | 2.9064 ± 0.4137 |
| Swin-Tiny | MAB-TS | 896 | 0.4790 ± 0.0272 | -0.0198 ± 0.0263 | 0.8553 ± 0.0077 | -0.0134 ± 0.0072 | 0.9810 ± 0.0025 | 3.1410 ± 0.2702 |
| Swin-Tiny | MAB-UCB | 896 | 0.4649 ± 0.0560 | -0.0339 ± 0.0492 | 0.8544 ± 0.0156 | -0.0143 ± 0.0170 | 0.9827 ± 0.0037 | 3.1671 ± 0.3974 |
| ViT-B/16 | Budgeted greedy | 896 | 0.6985 ± 0.0698 | 0.1384 ± 0.0586 | 0.8226 ± 0.0242 | -0.0302 ± 0.0212 | 0.9671 ± 0.0062 | 3.1827 |
| ViT-B/16 | MAB-TS | 896 | 0.6291 ± 0.0455 | 0.0690 ± 0.0337 | 0.8323 ± 0.0133 | -0.0205 ± 0.0079 | 0.9726 ± 0.0057 | 3.1827 |
| ViT-B/16 | MAB-UCB | 896 | 0.6638 ± 0.0782 | 0.1037 ± 0.0711 | 0.8226 ± 0.0167 | -0.0302 ± 0.0131 | 0.9709 ± 0.0086 | 3.1827 |

## Appendix B. Statistical and selection-cost details

Appendix B reports selection-cost accounting, the paired statistical summary from Section 5.7, and diagnostic figures for evaluation budgets, significance counts, and sensitivity cases.

*Appendix Table B1. Representative selection-cost and fairness summary for All-LLM 10% head-pruning rows.*

| Method | Candidate evals | Forward batches | Selection time (s) | Total time (s) | Mean total units |
|---|---|---|---|---|---|
| Random | 0 | 0.0 | 0.00 | 9.92 | 310.1 |
| Magnitude | 0 | 0.0 | 0.05 | 6.69 | 310.1 |
| Wanda-style | 0 | 69.9 | 0.96 | 6.91 | 290.5 |
| Taylor | 0 | 14.3 | 1.00 | 6.86 | 290.5 |
| Fisher | 0 | 14.3 | 0.98 | 6.80 | 290.5 |
| Budgeted greedy | 1851 | 7403.5 | 80.70 | 87.06 | 310.1 |
| MAB-UCB | 1851 | 7403.5 | 77.82 | 84.26 | 310.1 |
| MAB-TS | 1851 | 7403.5 | 76.32 | 82.54 | 310.1 |

Table B1 separates calibration use from counted paired candidate-evaluation trials. Random and magnitude selection do not perform candidate testing; Wanda-style, Taylor and Fisher use calibration information but are not adaptive paired candidate-evaluation procedures; budgeted greedy and MAB methods use the same counted candidate-evaluation convention.

Appendix Table B2 reports the statistical summary from Section 5.7 and identifies the 116-test family used for Benjamini-Hochberg correction.

*Appendix Table B2. Summary of the 28 highlighted paired language-model comparisons against budgeted greedy. Benjamini-Hochberg q-values use the full 116-test family.*

| Metric | Value |
|---|---|
| Highlighted MAB vs greedy rows | 28 |
| Bootstrap 95% CI excludes zero | 23/28 |

| Metric | Value |
|---|---|
| Paired p < 0.05 | 11/28 |
| BH q < 0.05 (116-test family) | 6/28 |

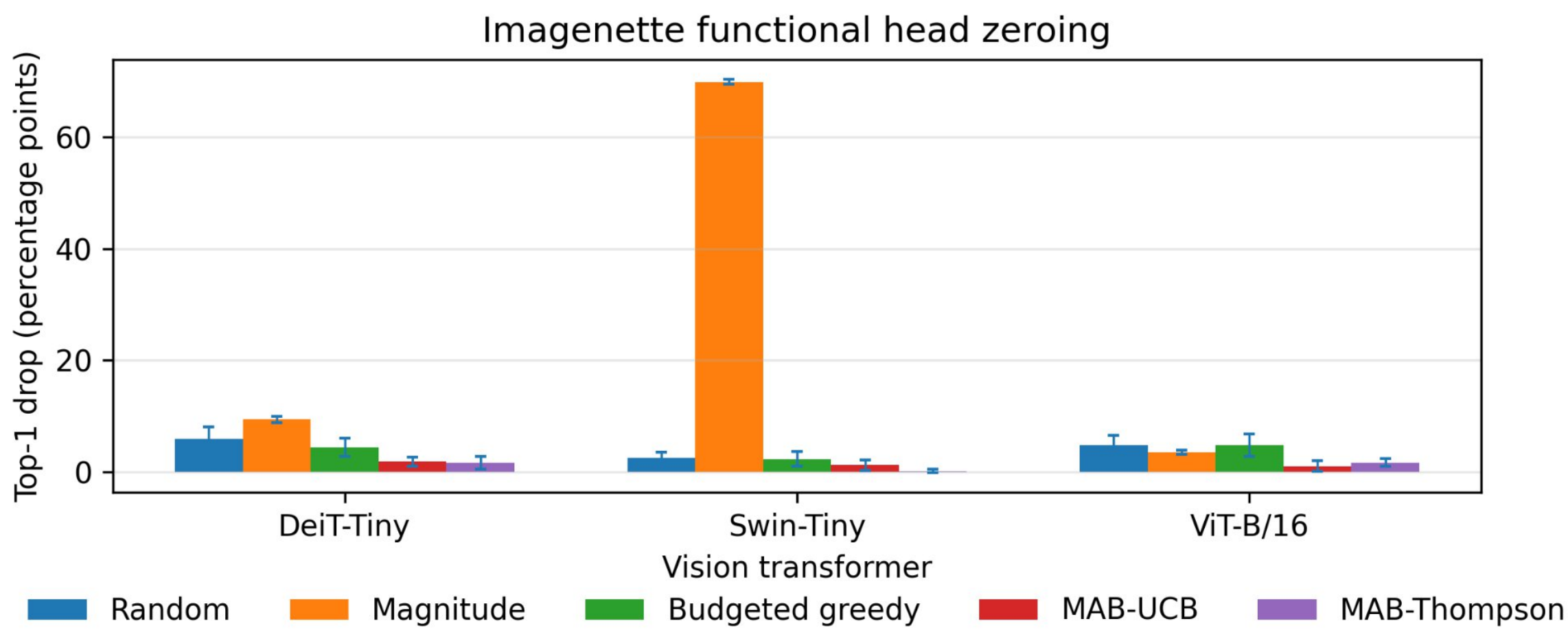

Appendix Figure B1. Full Imagenette all-method Top-1 drop after functional head zeroing. Lower Top-1 drop is better. This appendix figure includes Random, Magnitude, Budgeted greedy, MAB-UCB, and MAB-TS.

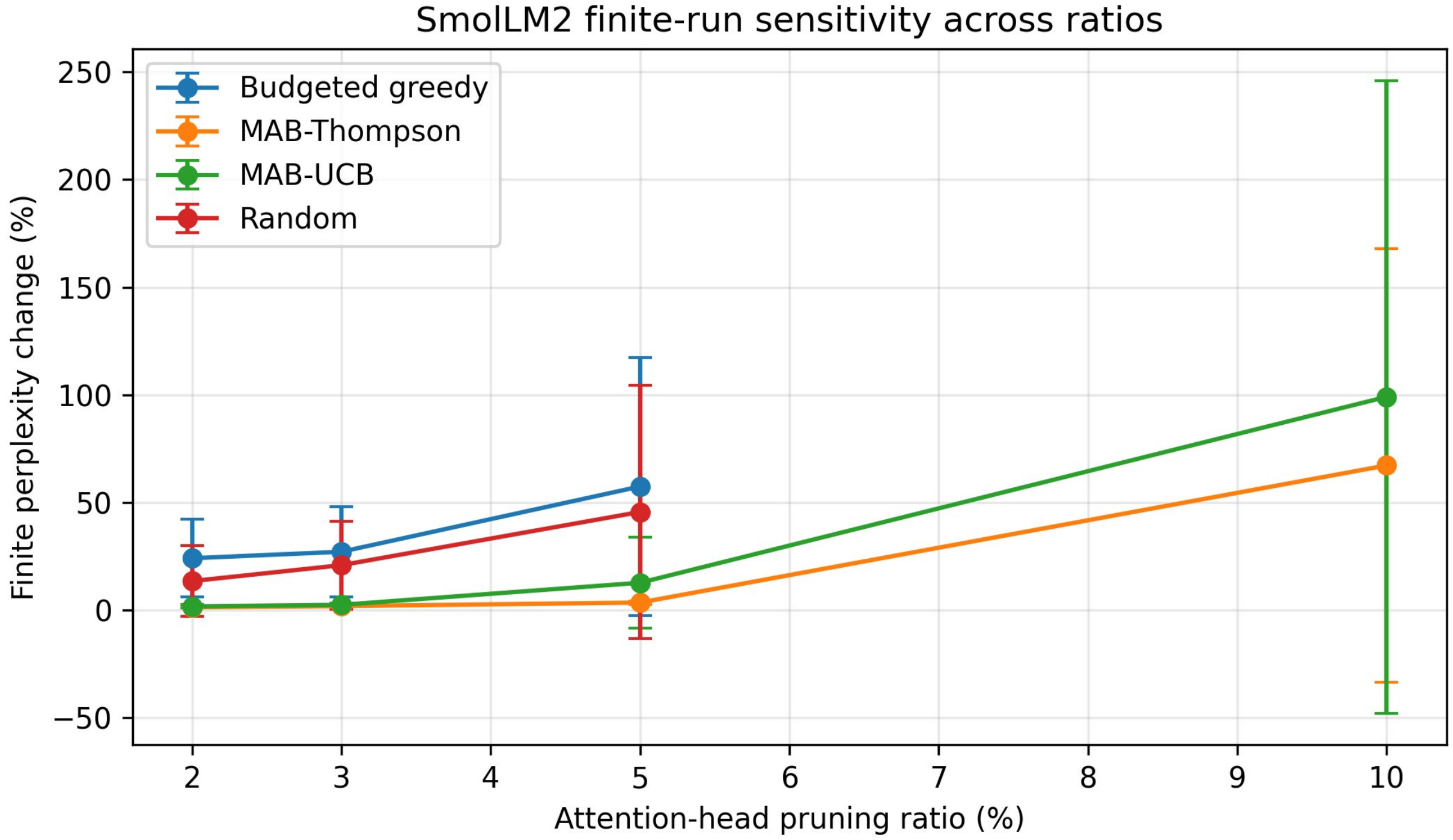

Appendix Figure B2. SmolLM2-360M finite-run sensitivity across attention-head pruning ratios. Values are based on Table 10, with non-finite runs excluded from finite means and retained as failures in the table.